\documentclass[10pt,twocolumn,letterpaper]{article}

\usepackage[pagenumbers]{cvpr} 
\usepackage{multirow}
\usepackage{float}
\usepackage{makecell}
\definecolor{cvprblue}{rgb}{0.21,0.49,0.74}
\usepackage[pagebackref,breaklinks,colorlinks,allcolors=cvprblue]{hyperref}

\def\paperID{***} 
\def\confName{CVPR}
\def\confYear{2026}

\title{MixerCSeg: An Efficient Mixer Architecture for Crack Segmentation \\ via Decoupled Mamba Attention}

\author{%
Zilong Zhao$^{1}$, 
Zhengming Ding$^{2}$, 
Pei Niu$^{1}$, 
Wenhao Sun$^{1}$,
Feng Guo$^{1,}$\thanks{Corresponding author}\\[1ex]
$^{1}$School of Qilu Transportation, Shandong University, China\\
$^{2}$Department of Computer Science, Tulane University,  USA\\
} 

\begin{document}
\maketitle
\begin{abstract}
Feature encoders play a key role in pixel-level crack segmentation by shaping the representation of fine textures and thin structures. Existing CNN-, Transformer-, and Mamba-based models each capture only part of the required spatial or structural information, leaving clear gaps in modeling complex crack patterns. To address this, we present MixerCSeg, a mixer architecture designed like a coordinated team of specialists, where CNN-like pathways focus on local textures, Transformer-style paths capture global dependencies, and Mamba-inspired flows model sequential context within a single encoder. At the core of MixerCSeg is the TransMixer, which explores Mamba’s latent attention behavior while establishing dedicated pathways that naturally express both locality and global awareness. To further enhance structural fidelity, we introduce a spatial block processing strategy and a Direction-guided Edge Gated Convolution (DEGConv) that strengthens edge sensitivity under irregular crack geometries with minimal computational overhead. A Spatial Refinement Multi-Level Fusion (SRF) module is then employed to refine multi-scale details without increasing complexity. Extensive experiments on multiple crack segmentation benchmarks show that MixerCSeg achieves state-of-the-art performance with only 2.05 GFLOPs and 2.54 M parameters, demonstrating both efficiency and strong representational capability. 
The code is available at \url{https://github.com/spiderforest/MixerCSeg}.
\end{abstract}

\section{Introduction}
\label{sec:intro}

As infrastructure such as roads and bridges ages and deteriorates over time, the issue of cracks has become increasingly prominent. As a key technology for monitoring and maintaining road health, road crack segmentation enables the timely detection of potential crack threats. However, achieving high-precision pixel-level crack segmentation remains a significant challenge due to the considerable morphological diversity, uneven texture distribution, and low contrast between cracks and their background \cite{pavement}. To address this challenge, existing crack segmentation models are mainly based on three architectures: Convolutional Neural Networks (CNNs), Transformers \cite{attention}, and Mamba \cite{mamba}.

\begin{figure}
  \centering
  \includegraphics[width=0.95\linewidth]{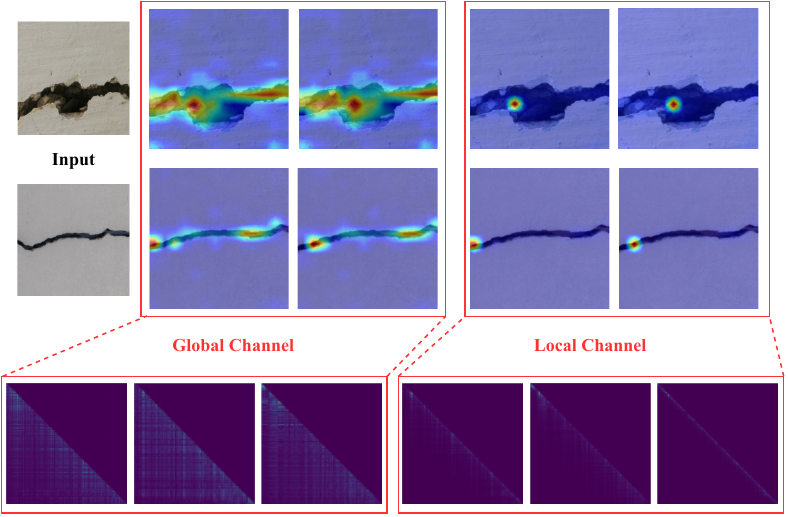}
  \vspace{-3mm}
    \caption{{Visualization results of heatmaps and attention maps in VMamba \cite{vmamba}. Benefiting from the multi-directional scanning mechanism, tokens in global channels can achieve a receptive field that covers the entire scope, rather than being limited to historical context. In contrast, local channels focus on the field of view of neighboring regions.}}
    \label{fig:glo_loc}
      \vspace{-5mm}
\end{figure}

Due to the strong capability in extracting local features and computational efficiency advantages, researchers have proposed various CNN-based crack segmentation models \cite{unet, carnet, rindnet}. However, their limited receptive fields make it difficult to effectively model long-range pixel dependencies, resulting in suboptimal performance when dealing with cracks of complex morphology.
Vision Transformer \cite{vit, swint} successfully introduced the Transformer architecture into the field of vision tasks. By leveraging attention mechanisms to explicitly model long-range dependencies between pixels, it has enabled crack segmentation models such as Crackmer \cite{crackmer} and DTrCNet \cite{dtrcnet} that outperform CNN-based approaches. However, this also results in increased computational complexity and reduced inference efficiency.
In recent years, the Mamba \cite{mamba} architecture has been introduced as a global attention mechanism with linear computational complexity. 
SCSegamba \cite{scsegamba} has designed a Structure-Aware Scanning Strategy to enhance crack segmentation performance. 
Nevertheless, this progressive data processing mechanism limits its ability to capture and utilize global context in a single forward pass \cite{mambavision}.

Recently, hybrid models with diverse architectures have been receiving increasing attention. MambaVision \cite{mambavision} adapts the Mamba structure and, for the first time, proposes a hybrid Mamba-Transformer architecture for visual tasks. VAMBA \cite{vamba}, targeting long-video understanding, constructs a hybrid model by integrating self-attention, cross-attention, and Mamba modules. RestorMixer \cite{restormixer} improves upon Mamba and Self-Attention \cite{attention} by proposing a hybrid model that combines CNN, Transformer, and Mamba to enhance image restoration performance. Although these approaches demonstrate the potential for performance improvement, they simply stack these components in a sequential or parallel manner without thoroughly considering the inherent relationships and differences among the architectures. As a result, the full advantages of the diverse architectural designs are not fully leveraged \cite{mambavision,restormixer}.

To address this challenge, we explored the potential attention mechanism of Mamba and proposed a novel hybrid architecture model called MixerCSeg for road crack segmentation. Specifically, we designed a Mixer architecture named TransMixer. It decouples tokens along the channel dimension into global tokens and local tokens based on the inherent attention characteristics of Mamba (as shown in Figure \ref{fig:glo_loc}). For the global tokens, self-attention is introduced to further model long-range dependencies. For the local tokens, a Local Refinement Module is designed to enhance fine-grained feature details through convolutional operations.
Additionally, to achieve pixel-level crack segmentation, we designed a Direction-guided Edge Gated Convolution (DEGConv). It employs spatial block processing and utilizes directional priors 
to enhance edge sensitivity under irregular crack geometries while minimizing computational overhead.
Finally, the Spatial Refinement multi-level feature fusion (SRF) module provides global contextual guidance for low-resolution features through cross-scale semantic interactions, generating high-precision segmentation results without incurring additional computational overhead.


In summary, our main contributions are as follows:
\begin{itemize}
    \item We introduce TransMixer, a newly designed feature encoding structure that assigns CNN, Transformer, and Mamba pathways distinct roles in capturing local textures, modeling global dependencies, and expressing contextual flow, forming a coordinated architecture rather than a stacked mixture of modules.

    \item We develop MixerCSeg, a crack segmentation model built on this encoder. It incorporates DEGConv to enhance direction-aware edge modeling and SRF to refine multi-scale spatial details with minimal computational cost.

    \item We conduct extensive experiments on multiple crack segmentation benchmarks and show that MixerCSeg achieves state-of-the-art performance with an efficient design of 2.05 GFLOPs and 2.54 M parameters.
\end{itemize}
\section{Related Work}
\label{sec:method}

\subsection{Road Crack Segmentation}
Current deep learning-based road crack segmentation methods can be primarily categorized into three types: CNN-based, Transformer-based, and Mamba-based approaches. Among CNN-based methods, DeepCrack \cite{deepcrack_nc} established a crack segmentation dataset and a benchmark segmentation model. SDDNet \cite{sddnet} employs separable convolutions to achieve efficient real-time crack segmentation. SFIAN \cite{sfian} enhances segmentation accuracy by designing a selective fusion and irregularity-aware network. BARNet \cite{barnet} introduces gradient priors and develops a crack segmentation model that integrates image features with gradient information. However, due to the limitations of CNNs in capturing global receptive fields, they struggle to effectively model long-range dependencies, which restricts further performance improvements in pixel-level crack segmentation.

ViT \cite{vit} successfully introduced the attention \cite{attention} mechanism into visual tasks. Leveraging the advantage of attention mechanisms in long-range dependency modeling, STA \cite{sta}  utilized Swin-Transformer \cite{swint} to extract slender crack features; Crackmer \cite{crackmer} designed a parallel segmentation architecture combining convolution and transformer to enhance the characterization of crack details; CrackFormer \cite{crackformer}proposed a scaling-attention block to suppress non-semantic features and sharpen semantic cracks. Despite these advancements, the quadratic computational complexity introduces new computational efficiency challenges.

The Mamba \cite{mamba} model, leveraging its advantage of linear computational complexity, has been successfully introduced into the visual domain by studies such as ViM \cite{vim} and VMamba \cite{vmamba}, providing new perspectives for road crack segmentation. MambaCrackNet \cite{mambacracknet} incorporates residual Mamba blocks to reduce background interference and improve the detection rate of fine cracks. CrackMamba \cite{crackmamba} and SCSegamba \cite{scsegamba} designed serpentine and structure-aware scanning strategies, respectively, enhancing the model's ability to characterize complex crack features.

\begin{figure*}[t]
  \centering
  \includegraphics[width=0.95\linewidth]{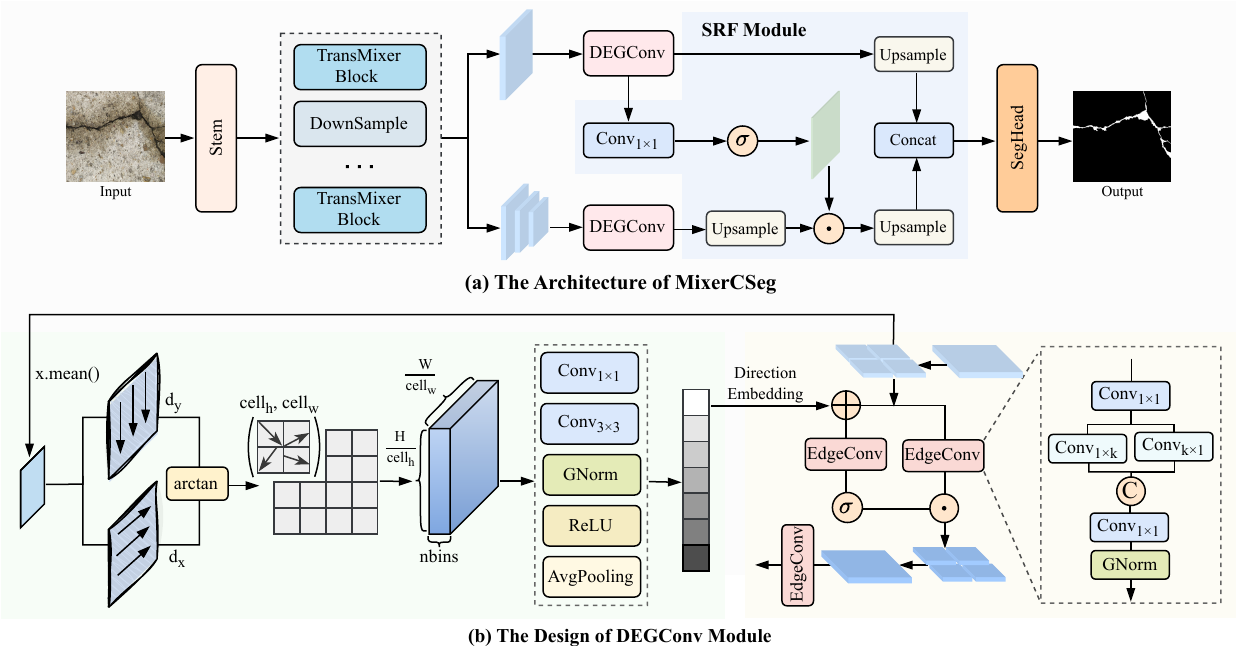}
  \vspace{-2mm}
    \caption{Overview of the proposed method. (a) illustrates the overall architecture of MixerCSeg. TransMixer blocks and downsampling operators are connected in series to extract multi-scale features. Before decoding, the DEGConv module is employed to enhance crack texture details. (b) demonstrates the design of the DEGConv, which integrates a spatial block strategy and directional prior knowledge.}
    \label{fig:overview}
      \vspace{-4mm}
\end{figure*}

\subsection{Hybrid Architecture Model}

Currently, common visual downstream tasks often adopt architectures such as Mamba-CNN or Transformer-CNN, while the Mamba-Transformer-CNN hybrid architecture remains in the exploratory stage. In the field of visual tasks, MambaVision \cite{mambavision} is the first to propose a backbone network based on a hybrid architecture, achieving performance surpassing that of ViT \cite{vit} and VMamba \cite{vmamba}. RestorMixer \cite{restormixer} is the first to introduce a hybrid model for image restoration tasks, aiming to leverage the strengths of different architectures. MambaFormer \cite{mambaformer} sequentially stacks Mamba and attention mechanisms to design a model for time series forecasting. VAMBA \cite{vamba} integrates cross-attention, self-attention, and Mamba to enhance performance in long video understanding tasks.

However, existing hybrid models often simply stack Mamba and Transformer modules without a refined design for the collaborative mechanism between different architectures. Based on an in-depth analysis of the hidden attention within Mamba, we propose a novel hybrid architecture. This architecture aims to leverage the unique advantages of each module at appropriate stages, thereby effectively enhancing the capability of crack feature extraction.

\section{Methodology}

\subsection{Preliminary: Hidden Attention of Mamba}
Given an input sequence $I =(x_1,x_2,...,x_L) \in \mathbb{R} ^{L \times d}$,  where $L$ denotes the token length and $d$ denotes the number of channels, the computation process of the Mamba block can be expressed as:
\begin{align}
Z = \sigma_1\left(\text{Linear}(I)\right)&; X = \text{Conv1D}(\text{Linear}(I)); \label{eq:1}\\
Y = &\text{SSM}(X); \label{eq:2} \\
O = \text{Lin}&\text{ear}(Y \odot  Z), \label{eq:3}
\end{align}
where $\sigma_1$ is the SiLU activation function, $\text{Linear}$ represents a regular linear projection, $\odot$ denotes element-wise multiplication, and $O
\in \mathbb{R}^{L \times d}$ is the output sequence. The discrete state transition equation and observation equation of the SSM are formulated as:
\begin{align}
    h_t = \bar A_th_{t-1} + \bar{B}_tx_t; \quad y_t = C_t h_t ,
\label{eq:ssm}
\end{align}
where $x_t$ is the input at time step t, $h_t$ is the hidden state, and $\bar{A}_t,\bar{B}_t$ and $C_t$ represent the decay factor of the hidden state, the update matrix controlling the hidden state, and the projection matrix for the output, respectively. They are computed as:
\begin{align}
\Delta_t=\text{Softp}&\text{lus}(x_t); \quad B_t, C_t = \text{Linear}_{B/C}(x_t); \\
\bar{A_t} &= \text{exp}(A\Delta_t); \quad \bar{B_t}=B_t\Delta_t,
\end{align}
where $A$ is a negative learnable matrix, $\Delta_t$ is a positive factor, ensuring the decay factor of the hidden state always satisfies $0<\bar A_t <1$ \cite{longmamba}. By expanding the recurrent computation steps in Eq. \ref{eq:ssm}, we quantify the contribution of the $j$-th token $x_j$ to the $i$-th token $x_i$ as:
\begin{align}
h_{i} &=\Sigma_{j=1}^{i}\left(\Pi_{k=j+1}^{i} \bar{A}_{k}\right) \odot \bar{B}_{j} \odot x_{j}; \\
y_{i} & =C_{i}^{T} \Sigma_{j=1}^{i}\left(\Pi_{k=j+1}^{i} \bar{A}_{k}\right) \odot \bar{B}_{j} \odot x_{j} \\
& =\Sigma_{j=1}^{i} \alpha_{i, j} \odot x_{j},
\end{align}
where $\alpha_{i, j} =C_{i}^{T}\left(\Pi_{k=j+1}^{i} \bar{A}_{k}\right) \odot \bar{B}_{j}$ is a weighting factor representing the contribution of the $j$-th token to the output of the $i$-th token, analogous to the attention scores in Transformers \cite{mambaatt, longmamba, decimamba}. Figure \ref{fig:glo_loc} presents the channel-wise visualization of this attention score. To investigate the fundamental factors influencing the attention score, we further expand the hidden state transition equation in Eq. \ref{eq:ssm}:
\begin{align}
    h_t = \bar A_th_{t-1} + \bar{B}_tx_t = e^{A\Delta_t}h_{t-1} + \Delta_tB_tx_t,
\label{eq:delta}
\end{align}
When $\Delta_t \to 0$, the current token is discarded and the token from time step $t-1$ is retained; when $\Delta_t>0$, the attenuated result of the token at time $t-1$ is added to the token at the current time step $t$. Therefore, $\Delta_t$ determines which tokens will influence future tokens \cite{decimamba}.

\begin{figure}[t]
  \centering
  \includegraphics[width=\linewidth]{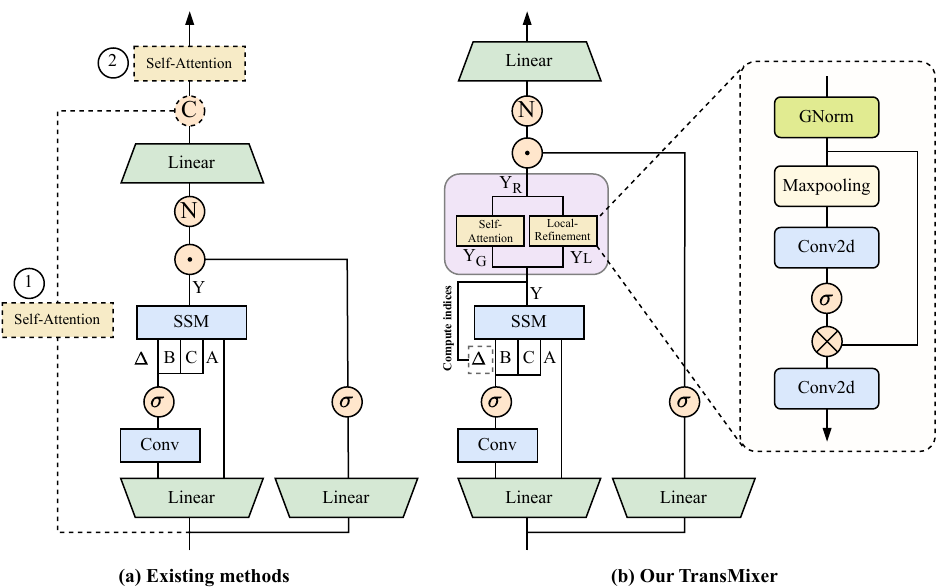}
    \caption{The difference between TransMixer and existing methods. (a) illustrates two common patterns in existing methods that hybridize Mamba blocks and Transformer blocks. (b) shows the design details of the TransMixer module, which decomposes features into global and local representations, enhancing them with Self-Attention and Local Refinement, respectively.}
      \vspace{-3mm}
    \label{fig:mixer}
\end{figure}

\subsection{Overview of Architecture}

Figure \ref{fig:overview} illustrates our proposed MixerCSeg model, which consists of three core components: the TransMixer Block, DEGConv module, and the SRF module. The TransMixer Block is designed to extract pixel-level crack features, DEGConv captures texture and topological cues of cracks, and SRF enables efficient multi-level feature fusion.

First, for an input image $P \in \mathbb{R}^{3\times H \times W}$, it is projected into visual features through a stem layer and processed by the TransMixer Block to obtain multi-scale feature maps $\{F_1, F_2, F_3, F_4\}$. These features are then fed into the DEGConv module, 
where a spatial block processing strategy is applied to introduce directional prior embeddings into the feature map, thereby enhancing the model's perception of texture and semantic features of cracks with diverse morphologies.
Finally, the SRF module integrates the multi-scale feature maps, and the segmentation head outputs a pixel-level crack segmentation result $r \in \mathbb{R}^{1\times H \times W}$.


\subsection{The Design of TransMixer}

The design intention of hybrid models is to combine the strengths of different architectures, enabling them to complement each other's advantages. Figure \ref{fig:mixer} (a) illustrates the two main implementation forms of current Transformer and Mamba hybrid models \cite{vamba, mambavision}, which integrate Mamba and the attention mechanism through parallel computation or sequential execution. However, both approaches share the same limitation: they merely stack modules in a simplistic manner without conducting an in-depth analysis of their core internal interaction logic, thereby restricting the full potential of hybrid architectures.

To address the limitations mentioned above, we conducted an in-depth analysis of the attention mechanism within the Mamba module and accordingly proposed the TransMixer module to more effectively capture visual features. The overall design of this module is illustrated in Figure \ref{fig:mixer} (b). According to Eq. \ref{eq:delta}, $\Delta_t$ determines the influence of historical tokens on the current token. Based on this, we sort $\Delta_t$ along the channel dimension, selecting the top $d_g = d \cdot \gamma$ tokens as global tokens and the remaining $d_l = d \cdot \left( 1 - \gamma\right)$ tokens as local tokens, with the corresponding index sets denoted as $g$ and $l$.

After the SSM execution is completed, we obtain the output $Y\in \mathbb{R}^{L \times d}$. Subsequently, using the indices $g$ and $l$, it is decomposed into a global representation $Y_G\in \mathbb{R}^{L \times d_g}$ and a local representation $Y_L\in \mathbb{R}^{L \times d_l}$. For the global representation $Y_G$, we feed it into a self-attention module to enhance the correlations among global tokens, resulting in the enhanced representation $Y_G^{\prime}$. For the local representation $Y_L$ we designed a simple Local Refinement module for processing, and its computational procedure is as follows:
\begin{align}
F_L \in \mathbb{R} ^{d_l \times H \times W} = \text{N}&\text{orm}(\text{Reshape}(Y_L));\\
F_L^{\prime} = \sigma_2(\text{Conv}_{1\times1}(\text{m}&\text{axpool2d}(F_l))) \odot F_L;\\
Y_l^{\prime} = \text{Flatt}&\text{en}(F_L^{\prime}),
\end{align}
where $\text{Norm}$ denotes normalization operation, and $\sigma_2$ is the sigmoid activation function.

Overall, the complete computational workflow of the TransMixer module is as follows: first, perform the operations described in Eq. \ref{eq:1} - Eq. \ref{eq:2} to obtain the output Y, and then execute: 
\begin{align}           
g,l &= f_{index}(\Delta_t) ; \\
Y_G, Y_L &= Y\left[:,g\right], Y\left[:,l\right]  ; \\
Y\left[:,g\right],Y\left[:,l\right] &= f_\text{self-att}(Y_G), f_\text{refine}(Y_L),
\end{align}
Finally, Eq. \ref{eq:3} is executed to complete the subsequent calculations, obtaining the feature map $F_i \in \mathbb{R}^{C_i\times H_i \times W_i}$.

\subsection{Direction-guided Edge Gated Convolution}
\label{sec:DEGConv}
Compared to common convolution that processes information passively, gated convolution can dynamically regulate the information flow, thereby more effectively preserving important features. In practical scenarios, cracks frequently extend into branches in multiple directions, posing significant challenges for models to accurately trace these intersecting and overlapping paths. To address this, we have designed a Direction-guided Edge Gated Convolution (DEGConv) module. This module integrates multi-perspective processing strategies with directional priors to precisely model complex crack structures, and its processing flow is illustrated in Figure \ref{fig:overview} (b).

\textbf{Rearrange.} For the feature map $F_i \in \mathbb{R}^{C_i\times H_i \times W_i}$ at the $i$-th layer, we partition it into $N$ non-overlapping views of size $h_i \times w_i$, denoted as $\mathcal{F}_{i} =\{F_i^1, F_i^2,...,F_i^N\}$, where $N = \frac{H_i}{h_i} \times \frac{W_i}{w_i}$. $F_i^j \in \mathbb{R}^{ C_i \times h_i \times w_i}$ represents the $j$-th local view feature map.

\textbf{Direction Embedding Generation.} We first average the input $F_i^j$ along the channel dimension to obtain $\tilde{F}_i^j \in \mathbb{R}^{1 \times h_i \times w_i}$ . Subsequently, the horizontal and vertical gradients $d_x$ and $d_y$ of this view are computed using the Sobel operator. Then, the radian value for each pixel is calculated via the arctangent function $\theta = \text{arctan}(\frac{d_y}{d_x})\in \mathbb{R}^{1 \times h_i \times w_i}$, and the result is constrained to the interval $\left[0,\pi \right]$.

To obtain a more compact feature representation and reduce interference from individual anomalous pixels, we further divide the view into cells of size $(cell_h , cell_w)$ and uniformly partition the radian interval $\left[0,\pi \right]$ into $n$ bins. For each cell $C_{a, b} (a=1,...,\frac{h_i}{cell_h};b=1,...,\frac{w_i}{cell_w})$, we calculate the bin index corresponding to the radian direction of each pixel, count the number of pixels within each bin, and weight them using the central angle value of the bin to obtain the orientation histogram feature of the cell. This process can be expressed as:
\begin{align}
\text{index}_{bin}&(u, v) = \left\lfloor \frac{\theta(u, v)}{\pi/n} \right\rfloor \\
\operatorname{hist}_{a, b}(k) =  &\sum_{(u, v) \in C_{a, b}} \mathbb{I}[\text{index}_{bin}(u, v) = k] \\
p_{a, b,k} &= c_{k} \cdot \frac{\operatorname{hist}_{a, b}(k)}{|C_{a, b}|}
\end{align}
where $\theta(u, v) \in \left[0,\pi\right]$ represents the directional radian value at position $(u, v)$, $n$ is the number of bins, and $k=0,...,n-1$ is the bin index. $\mathbb{I}\left[\cdot\right]$ denotes the indicator function. $c_k=\frac{\pi}{2n} + k \cdot \frac{\pi}{n}$ is the central radian value of the $k$-th bin, and $|C_{a, b}|=cell_h \times cell_w$ is the total number of pixels in the cell. Finally, we aggregate the features of all cells to obtain the spatial direction prior vector ${p} \in \mathbb{R}^{nbins \times \frac{h_i}{cell_h} \times \frac{w_i}{cell_w}}$, which is then transformed into a one-dimensional direction embedding vector:
\begin{align}
&f_p = \text{Conv}_{3\times3} \left(\text{Conv}_{1\times1}\left(p\right)\right); \\
\epsilon = \text{a}&\text{vgpool}(\text{Norm}(\text{ReLU}(f_p))) \in \mathbb{R}^{C_i}
\end{align}
Here, $\text{avgpool}$ denotes the adaptive average pooling operation, and $\epsilon$ is the resulting directional embedding vector.

\textbf{Edge Convolution.} To enhance the model's perception of crack texture features, we designed a lightweight edge convolution. First, we employ point-wise convolution to map the features into a low-dimensional space, thereby decreasing computational complexity. Subsequently, strip convolutions with kernel sizes of $1\times k$ and $k \times1$ are used to extract the directional features of cracks along the horizontal and vertical dimensions, respectively. Finally, the features from both directions are concatenated along the channel dimension, and the output result is obtained using depth-wise convolution.

\textbf{Gating mechanism.}
As shown in Figure \ref{fig:overview} (b), we first add the directional embedding features to the original feature maps, use EdgeConv and a sigmoid function to generate gating weights, and simultaneously apply EdgeConv to the original features, then combine them through element-wise multiplication:
\begin{align}
g = \sigma_2\left(\text{EdgeConv}(F_i^j + \epsilon ) \right); \\
{F_i^j}^{\prime} = g \odot \text{EdgeConv}(F_i^j)
\end{align}

Finally, the independent features from different spatial blocks are rearranged to restore the original spatial structure of the image. To further mitigate potential boundary misalignment issues introduced by the rearrangement, we introduce an additional EdgeConv layer for post-processing, ultimately obtaining the optimized feature representation $F_i^{\prime}$.

\begin{table*}[t]
\caption{Performance comparison on four crack datasets. The best result for each metric is in bold. 
}
\resizebox{\linewidth}{!}{
    \centering
    \renewcommand{\arraystretch}{1.2} 
    \begin{tabular}{c|cccc|cccc|cccc|cccc}
\Xhline{1pt}
    \multirow{2}{*}{Method}  &
    \multicolumn{4}{c}{DeepCrack} & \multicolumn{4}{c}{CamCrack789} & \multicolumn{4}{c}{CrackMap} & \multicolumn{4}{c}{Crack500}  \\
    \cline{2-17}
    &mIoU&ODS&OIS&F1 &mIoU&ODS&OIS&F1 &mIoU&ODS&OIS&F1 &mIoU&ODS&OIS&F1 \\
    \hline
    
    U-Net &0.8987&0.8868&0.8940&0.8999&\underline{0.8372}&\underline{0.7991}&\underline{0.8066}&0.7353&0.7983&0.7540&\textbf{0.7865}&0.7363&0.7105&0.6238&0.6610&0.5410 \\
    
    CarNet &0.8875&0.8688&0.8702&0.8939&0.8257&0.7830&0.7853& 0.7251&0.7568&0.6840&0.7117&0.6400&0.6428&0.5487&0.5921&0.3636\\

    RINDNet &0.8391&0.8087&0.8267&0.8377&0.8137&0.7628&0.7728&0.6050&0.7425&0.6745&0.6943&0.6699&0.7381&0.6469&0.6483&0.7119\\
    
    DTrCNet &0.8661&0.8473&0.8512&0.8566&0.8150&0.7719&0.7749&0.7836&0.7812&0.7328&0.7413&0.7276&0.7627&0.7012&0.7241&{0.7357}\\

    SCSegamba &\underline{0.9022}&\underline{0.8938}&0.8990&\underline{0.9110}&0.8268&0.7919&0.7939&\underline{0.7876}&\underline{0.8094}&\underline{0.7741} &0.7766&\underline{0.7678}&\underline{0.7778}&\underline{0.7244}&\underline{0.7370}&\underline{0.7553}\\

    RestorMixer &0.9008&0.8889&0.8957&0.9051&0.8356&0.7965&0.8018&0.7392&0.7887&0.7414&\underline{0.7844}&0.6631&0.7425&0.6637&0.6821&0.6962 \\
    
    MambaVision &0.8991&0.8851&\underline{0.9047}&0.8799&0.8146&0.7603&0.7626&0.7076&0.7737&0.7162&0.7657&0.6862&0.7015&0.6304&0.6480&0.4603\\

    \hline
    \textbf{MixerCSeg (Ours)}&\textbf{0.9151}&\textbf{0.9094}&\textbf{0.9197}&\textbf{0.9205}&\textbf{0.8409}&\textbf{0.8115}&\textbf{0.8202}&\textbf{0.8244}&\textbf{0.8123}&\textbf{0.7781}&0.7806&\textbf{0.7817}&\textbf{0.7824}&\textbf{0.7281}&\textbf{0.7483}&\textbf{0.7755}    \\
\Xhline{1pt}
    \end{tabular}%
    }
\label{tab:compare}
\end{table*}

\subsection{Spatial Refinement Multi-Level Feature Fusion}
Directly upsampling and fusing multi-level features not only causes misalignment of segmentation boundaries due to spatial inconsistency, but also fails to fully utilize the fine-grained spatial information present in high-resolution features. To address this, we designed a Spatial Refinement Multi-level Feature Fusion module (SRF). Guided by the rich spatial details of the high-resolution feature $F_1^{\prime}$, this module progressively upsamples and fuses the lower-level features $\{F_2^{\prime},F_3^{\prime},F_4^{\prime}\}$, thereby injecting fine-grained information into the semantic features and improving the accuracy of segmentation boundaries.

To leverage the detailed information from $F_1^{\prime}$ for guiding the fusion process, we first generate a spatial attention map from it. Meanwhile, the low-resolution features $\{F_2^{\prime},F_3^{\prime},F_4^{\prime}\}$ are upsampled to the same spatial dimensions as $F_1^{\prime}$. Subsequently, the attention map is used to perform a spatially weighted refinement on each upsampled feature, formulated as:
\begin{align}
\alpha &= \sigma_2\left(\text{Conv}_{1\times1}\left(F_i^{\prime}\right)\right);\\
F_i^{up} &= \text{Upsample}\left(F_i^{\prime}, \left(H_1,W_1\right)\right);\\
F_i^{\prime \prime}& = \alpha \odot F_i^{up},
\end{align}
where $\alpha$ denotes the spatial attention map, and $i \in \{2,3,4\}$. Next, we upsample all feature maps to the input image size, obtaining $\{F_1^{up};F_2^{up};F_3^{up};F_4^{up}\}$. These features are then concatenated along the channel dimension and fed into the segmentation head to predict pixel-level crack segmentation results. This process can be expressed as:
\begin{align}
    r = \mu\left([F_1^{up};F_2^{up};F_3^{up};F_4^{up}]\right) \in \mathbb{R}^{1\times H \times W},
\end{align}
where $\mu$ represents the segmentation head, which consists of a $1\times1$ convolution and an MLP.

\subsection{Loss Function}
Following \cite{scsegamba}, we employ a mixed loss function combining Binary Cross-Entropy (BCE) and Dice loss, with the ratio set to 1:5.

\section{Experiments}
\label{sec:experiment}

\subsection{Datasets and Metrics.}

\textbf{Datasets.} Our experiments were conducted on four crack benchmark datasets: DeepCrack \cite{deepcrack_nc}, Crack500 \cite{crack500}, CamCrack789 \cite{camcrack789}, and CrackMap \cite{crackmap}, containing 537, 3368, 789, and 120 crack images, respectively. These datasets encompass cracks in various scenarios, including walls, asphalt pavements, and cement surfaces, with diverse morphologies and scales. During the data processing stage, all images were resized to 512 $\times$ 512 pixels and randomly split into training, validation, and test sets in a 7:1:2 ratio.

\noindent \textbf{Metrics.} To evaluate the model performance, we adopted widely-used evaluation metrics, including the F1 score, Optimal Dataset Scale (ODS), Optimal Image Scale (OIS), and mean Intersection over Union (mIoU).

\subsection{Implementation Details.} 
\label{sec:implementation}
All experiments are conducted on a single NVIDIA A100 GPU. The model is trained for 50 epochs with a batch size of 1, optimized using the AdamW optimizer with an initial learning rate of 5e-4. For parameter settings, the hyperparameter $\gamma$ in the TransMixer block is set to 0.5 by default, and the cell size in DEGConv is configured to $8\times 8$. 
Specifically, the number of intervals n is set to 180 for the DeepCrack, CrackMap, and CamCrack789 datasets. In contrast, for the Crack500 dataset, where the primary challenge lies in interference from high background noise and the cracks exhibit smooth curvature and large width, n is set to 36.

\begin{table}[t]
\caption{Comparison of computational cost.}
\label{tab:cost}
\resizebox{\linewidth}{!}{
    \centering
    \renewcommand{\arraystretch}{1.2}
    \begin{tabular}{c|c|c|c}
\Xhline{1pt}
    {Method}  & {FLOPs (G)} & {Params (M)} & Memory (MiB)  \\
    \hline
    U-Net &204.38 & 28.99&4394  \\
    CarNet &19.00 &4.89 & \underline{1824} \\
    RINDNet &695.77 &59.39 & 5392 \\
    DTrCNet &122.55 & 63.83& 4952  \\
    
    SCSegamba &\underline{18.16} &\underline{2.80}& 2206 \\
    
    RestorMixer &98.71 &3.19& 10384 \\
    MambaVision &642.86&13.57& 5222 \\

    \hline
    \textbf{MixerCSeg (Ours)}&\textbf{2.05}&\textbf{2.54}&\textbf{1190}  \\
\Xhline{1pt}
    \end{tabular}%
}
\vspace{-2mm}
\end{table}

\begin{figure*}[ht]
  \centering
  \includegraphics[width=\linewidth]{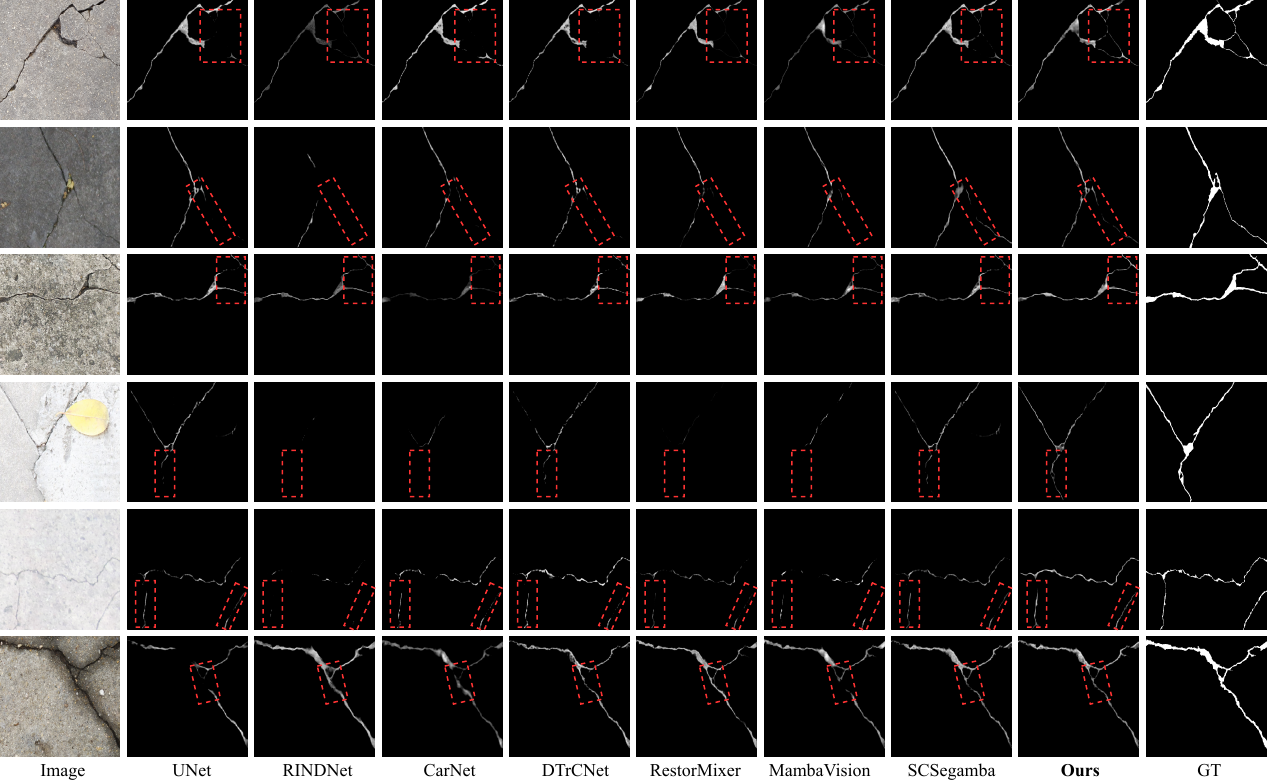}
    \caption{{Visualization results of MixerCSeg versus state-of-the-art methods in various environments.}}
    \label{fig:result_vis}
\end{figure*}

\subsection{Comparison with the State-of-the-arts}
\textbf{Quantitative Evaluation.} 
In Table \ref{tab:compare}, we compare the proposed MixerCSeg model with 7 SOTA models for image and crack segmentation tasks, including UNet \cite{unet}, CarNet \cite{carnet}, RINDNet \cite{rindnet}, DTrCNet \cite{dtrcnet},
RestorMixer \cite{restormixer}, SCSegamba \cite{scsegamba}, and MambaVision \cite{mambavision}. All comparative models were trained for 50 epochs with an input size of $512\times512$ to ensure experimental fairness. As shown in the table, our MixerCSeg achieves superior performance metrics across multiple datasets compared to existing advanced methods. Taking the DeepCrack \cite{deepcrack_nc} dataset as an example, MixerCSeg outperforms the second-best model SCSegamba \cite{scsegamba} by 1.43\% in mIoU and 1.04\% in F1-score, while achieving improvements of 1.78\% in mIoU and 4.61\% in F1-score compared to the hybrid architecture model MambaVision. These results fully demonstrate the effectiveness and superiority of our proposed method.

Furthermore, Table \ref{tab:cost} presents a comparison of computational efficiency between our proposed method and existing SOTA models. Our approach requires 2.54 M parameters and 2.05 GFLOPs, with a training memory usage of 1190 MiB. Compared to SCSegMamba, which is specifically designed for lightweight applications, our method reduces parameter count and computational cost by 9.3\% and 88.7\% respectively, while achieving a substantial memory reduction of 1016 MiB. Moreover, when compared to existing hybrid architectures MambaVision \cite{mambavision} and RestorMixer \cite{restormixer}, MixerCSeg demonstrates significantly lower computational resource requirements. These results fully validate the high efficiency and considerable potential of MixerCSeg for practical applications.

\noindent \textbf{Qualitative Evaluation.} To intuitively demonstrate the pixel-level segmentation results of MixerCSeg, we selected crack images from various complex environments in the test set for visualization, as shown in Figure \ref{fig:result_vis}. When facing challenging scenarios with variable sizes, diverse morphologies, and background noise interference, MixerCSeg can effectively capture crack regions.

\subsection{Ablation Study}
\textbf{Ablation study of components.} We use VMamba \cite{vmamba} as the encoder and integrate the Segformer \cite{segformer} decoder as the baseline model. The results of the ablation experiments are shown in Table \ref{tab:component}. We sequentially introduce the TransMixer, DEGConv, and SRF modules, and the crack segmentation performance is improved to varying degrees. This indicates that the TransMixer module can effectively extract image features, the DEGConv module helps capture the texture and morphological features of cracks, and the SRF module can generate accurate pixel-level crack representations. They work synergistically to achieve SOTA crack segmentation performance.

\begin{table*}[ht]
\caption{Component ablation experiments on the DeepCrack and CamCrack789 datasets.}
\resizebox{\linewidth}{!}{
    \centering
    \renewcommand{\arraystretch}{1.2}
    \begin{tabular}{ccc|ccc|cccc|cccc}
\Xhline{1pt}
    \multirow{2}{*}{TransMixer}  & \multirow{2}{*}{DEGConv} &\multirow{2}{*}{SRF} & \multirow{2}{*}{FLOPs (G)} & \multirow{2}{*}{Params (M)} & \multirow{2}{*}{Memory (MiB)} &
    \multicolumn{4}{c}{DeepCrack} & \multicolumn{4}{c}{CamCrack789}  \\
    \cline{7-14}
    & & &&&  &mIoU& ODS &OIS &F1  &mIoU& ODS &OIS &F1\\
    \hline
    
    & & &\underline{17.74} &\textbf{0.76} &\underline{1986} &0.8826&0.8696&0.8753&0.8867 &0.8283&0.7928&0.8007&0.8051  \\
    
    \checkmark & & &19.10 &\underline{2.50} &3594 &0.9016&0.8931&0.9045&0.9092&0.8359&0.8041&\underline{0.8164}&0.8163 \\

    \checkmark & \checkmark & &19.18&2.64 &3626 &\underline{0.9097}&\underline{0.9032}&\underline{0.9127}&\underline{0.9107}&\underline{0.8381}&\underline{0.8073}&0.8147&\underline{0.8211} \\

    \checkmark & \checkmark & \checkmark  &\textbf{2.05}& 2.54 &\textbf{1190} &\textbf{0.9151}&\textbf{0.9094}&\textbf{0.9197}&\textbf{0.9205}  &\textbf{0.8409}&\textbf{0.8115}&\textbf{0.8202}&\textbf{0.8244} \\
\Xhline{1pt}
    \end{tabular}%
    }
\label{tab:component}
\end{table*}

\begin{table}[ht]
\caption{Ablation experiments of different hybrid blocks on the DeepCrack and CamCrack789 datasets.}
\resizebox{\linewidth}{!}{
    \centering
    \renewcommand{\arraystretch}{1.2}
    \begin{tabular}{c|cccc|cccc}
\Xhline{1pt}
    \multirow{2}{*}{Encoder}  &
    \multicolumn{4}{c}{DeepCrack} & \multicolumn{4}{c}{CamCrack789}  \\
    \cline{2-9}
      &mIoU& ODS &OIS &F1  &mIoU& ODS &OIS &F1\\
    \hline
    
    MambaVision  &0.9020&0.8936&0.8979&0.9081&0.8307&0.7964&0.8132&0.8131\\

    RestorMixer  &\underline{0.9087}&\underline{0.9022}&\underline{0.9122}&0.9142&\underline{0.8373}&\underline{0.8063}&\textbf{0.8206}&\underline{0.8215} \\

    \textbf{TransMixer (Ours)}  &\textbf{0.9151}&\textbf{0.9094}&\textbf{0.9197}&\textbf{0.9205}   &\textbf{0.8409}&\textbf{0.8115}&\underline{0.8202}&\textbf{0.8244}\\
\Xhline{1pt}
    \end{tabular}%
    }
\label{tab:mixer}
\end{table}

\begin{table}[ht]
\caption{Analysis of Hyperparameter Experiments of the MixerCSeg Model on the DeepCrack and CamCrack789 Datasets.}
\resizebox{\linewidth}{!}{
    \centering
    \renewcommand{\arraystretch}{1.2}
    \begin{tabular}{c|cccc|cccc}
\Xhline{1pt}
    \multirow{2}{*}{Value} &
    \multicolumn{4}{c}{DeepCrack} & \multicolumn{4}{c}{CamCrack789}  \\
    \cline{2-9}
     & mIoU& ODS &OIS &F1  &mIoU& ODS &OIS &F1\\
    \hline

\hline
\multicolumn{9}{c}{ \textbf{$\gamma =0.5$, cell size is $(8,8)$ and $n=180$  }} \\
\hline
    - &\textbf{0.9151}&\textbf{0.9094}&\underline{0.9197}&\textbf{0.9205}   &\textbf{0.8409}&\textbf{0.8115}&{0.8202}&\textbf{0.8244}\\
\hline
\multicolumn{9}{c}{ \textbf{The ratio $\gamma$ of global channels}} \\
\hline
    0.3 &0.9098&0.9035&0.9117&0.9172 &0.8375&0.8065&\textbf{0.8240}&\underline{0.8243} \\

    0.7 &0.9133&0.9073&0.9182&0.9114 &0.8379&0.8072&0.8171&0.8231 \\

    0.9 &0.9084&0.9018&0.9112&0.9135 &0.8344&0.8026&0.8105&0.8166 \\

\hline
\multicolumn{9}{c}{\textbf{The cell size in the DEGConv module}} \\
\hline

   $\left(2,2\right)$ &0.9104&0.9040&0.9080&\underline{0.9187} &0.8374&0.8062&\underline{0.8213}&0.8230 \\
   $\left(4,4\right)$ &\underline{0.9141}&\underline{0.9081}&\textbf{0.9201}&0.9151 &0.8370&0.8059&0.8131&0.8210 \\

\hline
\multicolumn{9}{c}{\textbf{The number of bins $n$ in the DEGConv module}} \\
\hline
   $9$ &0.9067&0.8998&0.9106&0.9141 &0.8360&0.8042&0.8117&0.8175 \\
   $36$ &0.9125&0.9065&0.9173&0.9182 &0.8389&0.8088&0.8185&0.8217 \\
   $90$ &0.9136&0.9076&0.9191&0.9165 &\underline{0.8401}&\underline{0.8102}&0.8198&0.8240 \\
   $360$ &0.9077&0.9005&0.9044&0.9152 &0.8383&0.8072&0.8134&0.8220 \\
\Xhline{1pt}
    \end{tabular}%
    }
\label{tab:hyper}
\end{table}

In addition, after integrating the DEGConv module, the model only incurs an increase of 0.08 GFLOPs in computational complexity and 0.14 M in parameters.
When replacing the decoder of SegFormer with the SRF module, the computational complexity and GPU memory usage decrease significantly, and the parameters are also reduced by 0.1 M. These results fully demonstrate the lightweight advantages of the DEGConv and SRF modules.

\noindent \textbf{Effect of TransMixer.}
To validate the effectiveness of our hybrid architecture, we employed MambaVision \cite{mambavision} and RestorMixer \cite{restormixer} as encoders to extract multi-scale features, while keeping the other structures unchanged. The experimental results, as shown in Table \ref{fig:mixer}, demonstrate that TransMixer achieves superior segmentation performance, thereby verifying the effectiveness of our proposed method.

\noindent \textbf{The ratio $\gamma$ of global channels.}
By observing the attention weights in the channel dimension of the final layer of the Mamba module, we found that the number of channels with global attention is similar to that with only local attention.
Therefore, we set $\gamma$ to 0.5 by default. To validate the rationality of this setting, we conducted comparative experiments by setting $\gamma$ to 0.3, 0.7, and 0.9. The results, as shown in rows 6 - 8 of Table \ref{tab:hyper}, demonstrate that when $\gamma$ deviates from 0.5, the model performance declines to varying degrees, indicating that the default value of $\gamma = 0.5$ is a reasonable configuration.

\noindent \textbf{The cell size in the DEGConv module.} In crack images, the proportion of actual effective crack regions is low and background interference is significant. Smaller cell sizes are highly sensitive to noise, so we set the default cell size to $8\times8$. To verify the rationality of this setting, we additionally designed comparative experiments with two other cell sizes ($2\times2$ and $4\times4$). As shown in rows 10 - 11 of Table \ref{tab:hyper}, smaller cell sizes correspond to the lowest segmentation performance, which fully validates the rationality of our default setting.

\noindent \textbf{The number of bins $n$ in the DEGConv module. }
The number of bins directly affects the granularity of directional embedding, with its default set to 180. To validate the rationality of this configuration, we conducted a comparative experiment as shown in rows 13 - 16 of Table \ref{tab:hyper}. The results indicate that as the number of bins increases, model performance shows a slow upward trend, but when the number of bins becomes excessively large, performance declines instead. This is because overly fine-grained interval partitioning causes each bin to contain only a single value, which contradicts the original intention of this parameter setting.

\begin{table}[t]
\caption{Results of direction embeddings generated based on ARConv-predicted angles and those generated by the proposed method on the DeepCrack dataset.}
\label{tab:embedd}
    \centering
    \renewcommand{\arraystretch}{1.2}
    \begin{tabular}{c|c|c|c|c}
\Xhline{1pt}
    {Method}  & mIoU & ODS& OIS & F1  \\
    \hline
    ARConv \cite{arconv} &0.9106 &0.9043&0.9093&0.9170  \\

\textbf{Ours}&\textbf{0.9151}&\textbf{0.9094}&\textbf{0.9197}&\textbf{0.9205}  \\
\Xhline{1pt}
    \end{tabular}%

\end{table}

\noindent \textbf{Effect of directional embeddings in the DEGConv module.}
The method of directly predicting angles and then computing directional embeddings \cite{arconv} is constrained by the lack of supervision, making it difficult to ensure the reliability of the directional embeddings. In contrast, we leverage directional priors to construct the embeddings directly, endowing the model with more reasonable geometric interpretability. The results in Table \ref{tab:embedd} confirm the effectiveness of this design.
\section{Conclusion}
In this paper, we propose a hybrid model called MixerCSeg for crack segmentation, which comprises three key components. First, the TransMixer module fully leverages the strengths of Mamba, Transformer, and CNN. It utilizes the implicit attention mechanism in Mamba to divide tokens into local and global tokens, naturally combining the Transformer and CNN frameworks to capture multi-scale crack features efficiently. Secondly, DEGConv enhances the representation of crack texture details and morphological characteristics at minimal cost through directional priors and a gating mechanism. Finally, the SRF module refines low-resolution features using high-resolution information to achieve pixel-level crack segmentation. Extensive experimental validation on crack datasets demonstrates that MixerCSeg achieves SOTA performance in crack segmentation while requiring lower computational resources.

\section{Acknowledgements}
This work is partially supported by the Natural Science Foundation of China (Grant Number: 52308457), the Natural Science Foundation of Shandong Province (Grant Number: ZR2023QE220), and the China Postdoctoral Science Foundation (Grant Number: 2024M761811).

{
    \small
    \bibliographystyle{ieeenat_fullname}
    \bibliography{main}
}

\clearpage
\maketitlesupplementary

\section{Additional Ablation Experiments}

\subsection{Ablation studies of the TransMixer Module}
To thoroughly investigate the effectiveness of the TransMixer module, we conducted comprehensive ablation experiments focusing on this component. Specifically, we evaluated the model performance under three conditions: removing the TransMixer module, removing global tokens, and removing local tokens. 
As shown in Table \ref{tab:mixer_component}, when the TransMixer module is removed, the model performance decreases by 0.76\% and 0.49\% on the DeepCrack \cite{deepcrack_nc} and CamCrack789 \cite{camcrack789} datasets, respectively. Similarly, models with either the local tokens or the global tokens removed also exhibit varying degrees of performance decline. These experimental results validate the effectiveness of the TransMixer module, demonstrating that each sub-component plays a specific and meaningful role.

Furthermore, we replaced the maxpooling with avgpooling in the Local Refinement module. As shown in Table \ref{tab:pooling}, compared to avgpooling, maxpooling is more effective in enhancing the most critical and salient crack features within local regions.

\begin{table}[ht]
\caption{Experimental results of decoding global tokens and local tokens in the TransMixer module on the DeepCrack and CamCrack789 datasets.}
\resizebox{\linewidth}{!}{
    \centering
    \renewcommand{\arraystretch}{1.2}
    \begin{tabular}{cc|cccc|cccc}
\Xhline{1pt}
    \multirow{2}{*}{Local}  & \multirow{2}{*}{Global} &
    \multicolumn{4}{c}{DeepCrack} & \multicolumn{4}{c}{CamCrack789}  \\
    \cline{3-10}
     & &mIoU& ODS &OIS &F1  &mIoU& ODS &OIS &F1\\
    \hline
    
    & &  0.9082&0.9009&0.9054&0.9155&0.8368&0.8057&0.8158&\underline{0.8252}\\

    \checkmark  &&0.9120&0.9057&0.9138&\textbf{0.9206}&0.8382&\underline{0.8081}&0.8179&{0.8247}\\

    & \checkmark &\underline{0.9136}&\underline{0.9079}&\underline{0.9169}&0.9204&\underline{0.8384}&{0.8076}&\underline{0.8200}&\textbf{0.8275} \\
    
    \checkmark & \checkmark &\textbf{0.9151}&\textbf{0.9094}&\textbf{0.9197}&\underline{0.9205}   &\textbf{0.8409}&\textbf{0.8115}&\textbf{0.8202}&{0.8244}\\

\Xhline{1pt}
    \end{tabular}%
    }
\label{tab:mixer_component}
\end{table}

\begin{table}[ht]
\caption{Experimental results of different pooling operations in the Local Refinement Module on the DeepCrack and CamCrack789 Datasets.}
\resizebox{\linewidth}{!}{
    \centering
    \renewcommand{\arraystretch}{1.2}
    \begin{tabular}{c|cccc|cccc}
\Xhline{1pt}
    \multirow{2}{*}{Methods} &
    \multicolumn{4}{c}{DeepCrack} & \multicolumn{4}{c}{CamCrack789}  \\
    \cline{2-9}
      &mIoU& ODS &OIS &F1  &mIoU& ODS &OIS &F1\\
    \hline
    
    Avgpooling&0.9147&0.9089&0.9192&0.9140&0.8399&0.8100&\textbf{0.8227}&\textbf{0.8273}\\
    \textbf{Maxpooling}&\textbf{0.9151}&\textbf{0.9094}&\textbf{0.9197}&\textbf{0.9205}   &\textbf{0.8409}&\textbf{0.8115}&{0.8202}&{0.8244}\\

\Xhline{1pt}
    \end{tabular}%
    }
\label{tab:pooling}
\end{table}

\subsection{Ablation studies of the DEGConv Module}


In Section \ref{sec:DEGConv}, the DEGConv module consists of four components: Rearrange, DEG (Directional Embedding Generation), Edge Convolution, and Gating. To systematically evaluate the necessity and contribution of each component, we conducted the experiments shown in Table \ref{tab:degconv}. The results demonstrate that each component contributes to the improvement of segmentation performance. In particular, the DEG operation leads to notable performance gains, achieving an mIoU increase of 0.36\% and 0.46\% on the DeepCrack \cite{deepcrack_nc} and CamCrack789 \cite{camcrack789} datasets, respectively. Although the Rearrange operation alone yields relatively modest improvements, with mIoU gains of 0.24\% and 0.05\% on the two datasets, respectively, it plays a critical role by optimizing the spatial arrangement of features through a spatial block approach. This results in a more structured feature input, which serves as a solid foundation for the subsequent DEG and Edge Convolution steps, thereby contributing to the overall performance enhancement of the module.

\begin{table}[ht]
\caption{The ablation experiment results of DEGConv on the DeepCrack and CamCrack789 datasets, including the Rearrange, DEG, Edge Convolution, and Gating operations.}
\resizebox{\linewidth}{!}{
    \centering
    \renewcommand{\arraystretch}{1.2}
    \begin{tabular}{cccc|cccc|cccc}
\Xhline{1pt}
    \multirow{2}{*}{Rearrange}  & \multirow{2}{*}{DEG} &\multirow{2}{*}{EdgeConv}  &\multirow{2}{*}{Gating} &\multicolumn{4}{c}{DeepCrack} & \multicolumn{4}{c}{CamCrack789}  \\
    \cline{5-12}
    & & &   &mIoU& ODS &OIS &F1  &mIoU& ODS &OIS &F1\\
    \hline
    
    & & & &0.9052&0.8980&0.9042&0.9052 &0.8334&0.8008&0.8122&0.8114  \\
    
    \checkmark & & & &0.9074&0.9004&0.9127&0.9130&0.8338&0.8009&{0.8079}&0.8162 \\

    \checkmark & \checkmark & &&{0.9107}&{0.9044}&\underline{0.9143}&\underline{0.9174}&{0.8376}&{0.8065}&\textbf{0.8237}&\underline{0.8252} \\

    \checkmark & \checkmark & \checkmark & &\underline{0.9132}&\underline{0.9072}&{0.9125}&\textbf{0.9205}  &\underline{0.8398}&\underline{0.8092}&{0.8193}&\textbf{0.8254} \\

    \checkmark & \checkmark & \checkmark & \checkmark&\textbf{0.9151}&\textbf{0.9094}&\textbf{0.9197}&\textbf{0.9205}  &\textbf{0.8409}&\textbf{0.8115}&\underline{0.8202}&{0.8244} \\
\Xhline{1pt}
    \end{tabular}%
    }
\label{tab:degconv}
\end{table}

\begin{table*}[ht]
\caption{Experiments on different layer depths on the DeepCrack and CamCrack789 Datasets.}
\resizebox{\linewidth}{!}{
    \centering
    \renewcommand{\arraystretch}{1.2}
    \begin{tabular}{c|ccc|cccc|cccc}
\Xhline{1pt}
    \multirow{2}{*}{Depth} & \multirow{2}{*}{FLOPs (G)} & \multirow{2}{*}{Params (M)} & \multirow{2}{*}{Memory (MiB)} &
    \multicolumn{4}{c}{DeepCrack} & \multicolumn{4}{c}{CamCrack789}  \\
    \cline{5-12}
    &&&  &mIoU& ODS &OIS &F1  &mIoU& ODS &OIS &F1\\
    \hline
    
    1&\textbf{2.05}&\textbf{2.54}&\textbf{1190}&\textbf{0.9151}&\textbf{0.9094}&\underline{0.9197}&\textbf{0.9205}   &\textbf{0.8409}&\textbf{0.8115}&{0.8202}&\underline{0.8244}   \\

    2&\underline{3.51}&\underline{4.76}&\underline{1550}&\underline{0.9141}&\underline{0.9084}&\textbf{0.9213}&0.9071&\underline{0.8381}&\underline{0.8069}&\underline{0.8222}&{0.8208}\\

    4&6.42&9.20&2080&0.9126&{0.9065}&{0.9166}&\underline{0.9173}&{0.8375}&{0.8065}&\textbf{0.8237}&\textbf{0.8252} \\
    
    6&9.33&13.63&3818&{0.9073}&{0.9004}&{0.9127}&0.9130&{0.8336}&{0.8005}&{0.8134}&{0.8203}\\


\Xhline{1pt}
    \end{tabular}%
    }
\label{tab:depth}
\end{table*}
\subsection{Impact of Layer Depth}

In the design of the MixerCSeg network architecture, the depth of each TransMixer block is set to 1 to balance computational resources and performance. We quantitatively analyzed the impact of block depth on model performance through controlled experiments in Table \ref{tab:depth}. While keeping other parameters constant, we tested configuration schemes with TransMixer block depths of 2, 4, and 6 layers, respectively. The experimental results show that when the depth of each layer is set to 1, the model demonstrates optimal segmentation performance with lower computational resource consumption. In comparison, when the network depth is increased to 2 layers, the model’s computational load is 3.51 GFLOPs, the number of parameters is 4.76 M, and the memory usage is 1550 MiB. These values represent an increase of 71.2\% in computation, 87.4\% in parameters, and 30.2\% in memory usage. However, since crack segmentation tasks are highly dependent on local fine-grained features such as pixel-level crack morphology and topological continuity, the finer details in deeper networks may gradually become blurred through multiple layers of propagation. This can lead to issues such as over-smoothing at edges. In addition, the difficulty of parameter optimization significantly increases, resulting in degraded performance. Therefore, setting the depth of each TransMixer block to 1 not only ensures the accuracy of crack segmentation but also effectively controls model complexity, providing feasibility for deployment on edge devices.

\subsection{Impact of the number of bins on the Crack500}
\label{sec:rationale}
In Section \ref{sec:implementation} of the experimental part, we mentioned that the number of bins $n$ was set to 36 rather than 180 on the Crack500 dataset. To systematically investigate the influence of this parameter, we conducted detailed ablation experiments on the Crack500 \cite{crack500} dataset, as shown in Table \ref{tab:nbins}, evaluating the model performance with nset to 9, 18, 36, 90, and 180, respectively. The experimental results indicate that the model achieves the best performance when $n = 36$.

\begin{figure}[H]
  \centering
  \includegraphics[width=\linewidth]{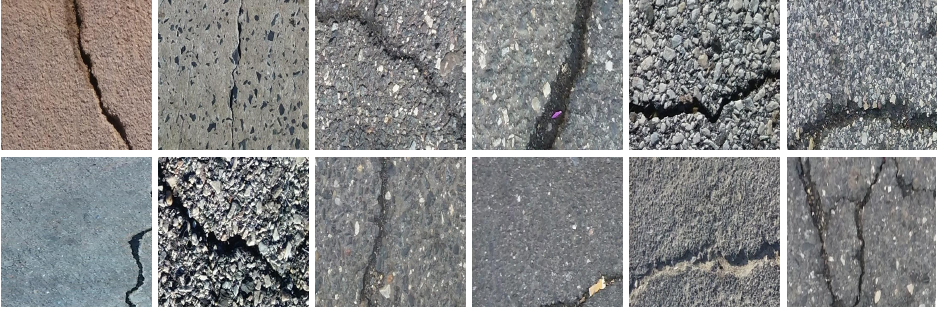}
  \vspace{-3mm}
    \caption{{Representative crack images from the Crack500 dataset.}}
    \label{fig:crack500}
      \vspace{-3mm}
\end{figure}

To further explore the intrinsic causes of this phenomenon, we conducted a detailed observation of the crack morphologies in the Crack500 dataset. 
As shown in Figure \ref{fig:crack500}, most cracks are characterized by large width and small curvature with significant background noise interference, while a few cracks (the last one) face the challenge of complex topological structures, and the ratio of the two types is close to 15:1.
This observation is consistent with the phenomenon that the optimal performance is achieved when $n=36$. A moderate $n$ can not only effectively capture the features of cracks with gentle curvature but also maintain good representation ability for wide cracks, and is also capable of handling a small proportion of morphologically complex cracks.

\begin{table}[H]
\caption{Ablation studies of the number of bins $n$ on the Crack500 dataset.}
\label{tab:nbins}
    \centering
    \renewcommand{\arraystretch}{1.2}
    \begin{tabular}{c|c|c|c|c}
\Xhline{1pt}
    {$n$}  & mIoU & ODS& OIS & F1  \\
    \hline
    9  &0.7805 &0.7269&\underline{0.7467}&0.7575  \\
    18  &\underline{0.7814} &\underline{0.7273}&0.7441&0.7631  \\
    \textbf{36} &\textbf{0.7824}&\textbf{0.7281}&\textbf{0.7483}&\textbf{0.7755}  \\
    90  &0.7813 &0.7264&0.7455&\underline{0.7692}  \\
    180  &0.7774 &0.7212&0.7409&0.7592  \\
\Xhline{1pt}
    \end{tabular}%

\end{table}

\section{Qualitative Analysis of SRF Module}
In the component ablation experiments presented in Table \ref{tab:component}, after replacing the decoder of SegFormer \cite{segformer} with the SRF module, the model's computational cost decreased by 89.3\%, the number of parameters was reduced by 33.8\%, and GPU memory usage dropped by 67.2\%. Meanwhile, on two crack segmentation benchmark datasets, DeepCrack \cite{deepcrack_nc} and CamCrack789 \cite{camcrack789}, the mIoU achieved improvements of 0.59\% and 0.32\%, respectively.

\begin{figure*}[ht]
  \centering
  \includegraphics[width=0.9\linewidth]{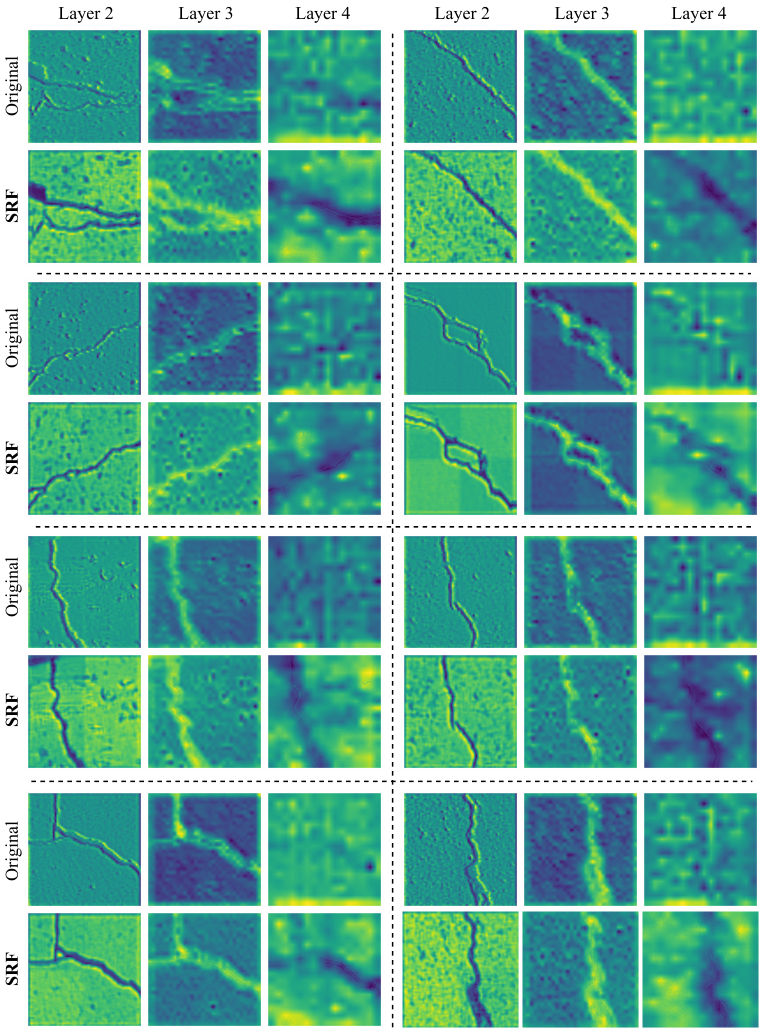}
  \vspace{-3mm}
    \caption{Visual comparison results of multi-scale features before and after SRF module processing. The distinction between cracks and the background is more prominent, and more notably, the crack features in the layer 4 have been significantly enhanced.}
    \label{fig:srf_feature}
      \vspace{-3mm}
\end{figure*}

To gain a deeper insight into the underlying mechanism of this performance gain, we conducted visual analyses on both the original and SRF-processed multi-scale feature maps. As illustrated in Figure \ref{fig:srf_feature}, after being processed by the SRF module, the semantic discriminability between crack regions and background in feature maps across various scales was significantly enhanced. The raw $F_4^{\prime}$ only exhibited weak texture responses; however, after refinement by $F_1^{\prime}$, the activation values in crack regions were substantially strengthened, forming clearer semantic boundaries. This improvement provides more discriminative cues for the segmentation head. Additionally, the alignment capability of the optimized features with high-resolution details was enhanced, effectively alleviating the issue of insufficient fusion between high-level semantics and low-level details in the decoder. Ultimately, these enhancements contributed to the improvement in segmentation accuracy.

%

\end{document}